\documentclass[11pt]{article}

\usepackage[preprint]{acl}

\usepackage{times}
\usepackage{latexsym}
\usepackage[utf8]{inputenc}
\usepackage{booktabs}
\usepackage{nicefrac} 
\usepackage{microtype} 
\usepackage[T1]{fontenc}
\usepackage{amsfonts} 
\usepackage[utf8]{inputenc}

\usepackage{microtype}

\usepackage{inconsolata}

\usepackage{graphicx}
\usepackage[table]{xcolor}         
\usepackage{multirow}
\usepackage{caption} 
\usepackage[most]{tcolorbox}
\tcbuselibrary{breakable,skins}
\newcommand{\equalcontrib}{\textsuperscript{*}}
\newtcolorbox{promptbox}[2][]{%
    enhanced,
    breakable,
    width=\linewidth,
    colback=gray!5,
    colframe=gray!30,
    sharp corners,
    boxrule=0.5pt,
    left=10pt, right=10pt, top=10pt, bottom=10pt,
    fontupper=\small\ttfamily\raggedright,
    before upper={\setlength{\parindent}{0pt}},
    title={#2},
    coltitle=black,
    colbacktitle=gray!15,
    fonttitle=\bfseries\small,
    boxed title style={
        boxrule=0pt,
        sharp corners
    },
    attach boxed title to top left={
        yshift=-2mm,
        xshift=2mm
    },
    segmentation style={draw=none},
    #1
}
\newcommand\blfootnote[1]{%
  \begingroup
  \renewcommand\thefootnote{}\footnote{#1}%
  \addtocounter{footnote}{-1}%
  \endgroup
}
\title{OphIn-500K: Curating Web-Scale Visual Instructions for Scaling Ophthalmic Multimodal Large Language Models}

\author{
  {\mdseries Xuanzhao Dong$^{1}$}\thanks{These authors contributed equally to this work.} \quad
  {\mdseries Wenhui Zhu$^{1\equalcontrib}$}\thanks{Now at LinkedIn.} \quad
  {\mdseries Xiwen Chen$^{2\equalcontrib}$}\thanks{Now at Morgan Stanley.} \quad
  {\mdseries Hao Wang$^{2\equalcontrib}$} \quad
  {\mdseries Xin Li$^{1\equalcontrib}$} \quad
  {\mdseries Yujian Xiong$^{1}$} \\
  {\mdseries Jiajun Cheng$^{1}$} \quad
  {\mdseries Jingjing Wang$^{2}$} \quad
  {\mdseries Xiaobing Yu$^{3}$} \quad
  {\mdseries Haiyu Wu$^{4}$} \quad
  {\mdseries Shao Tang$^{5}$} \quad
  {\mdseries Zhipeng Wang$^{6}$} \\
  {\mdseries Langechuan Liu$^{7}$} \quad
  {\mdseries Shan Lin$^{1}$} \quad
  {\mdseries Oana Dumitrascu$^{8}$} \quad
  {\mdseries Yalin Wang$^{1}$}\thanks{Corresponding author: \texttt{ylwang@asu.edu}} \\\\
  $^{1}$Arizona State University \quad
  $^{2}$Clemson University \quad
  $^{3}$Washington University in St. Louis \\
  $^{4}$University of Notre Dame \quad
  $^{5}$Florida State University \quad
  $^{6}$Rice University \\
  $^{7}$NVIDIA \quad
  $^{8}$Mayo Clinic \\
}

\begin{document}
\maketitle

\blfootnote{Preprint. Under review.}

\begin{abstract}
The advancement of general medical Multimodal Large Language Models (MLLMs) has shown great potential for building conversational assistants to support clinical diagnosis. However, their adaptation to highly specialized domains such as ophthalmology remains underexplored, primarily due to the scarcity of large-scale, domain-specific instruction-tuning data. Existing ophthalmic datasets for conversational agents are often limited in scale and largely rely on images from established public benchmarks, limiting the scalability of ophthalmic MLLMs and their ability to capture real-world clinical complexity. To address this gap, we propose \textbf{OphIn-Engine}, an ophthalmology-specific instruction data curation pipeline that constructs high-quality instruction data from open-access ophthalmology web-scale videos. The pipeline integrates multimodal transcription for extracting image-transcript pairs, visual cue separation and scoring for identifying clinically relevant visual descriptions, and instruction synthesis with quality control for generating accurate and diverse clinical dialogues. Using this engine, we introduce \textbf{OphIn-500K}, a large-scale multimodal ophthalmology instruction-tuning dataset containing over 500,000 instruction instances and more than 151,000 unique images from over 29,000 video clips, formatted as visual question answering (VQA), multi-turn conversational interactions, and chain-of-thought (CoT) reasoning. Built upon this dataset, we further develop \textbf{OphIn-VL}, an ophthalmology-specific MLLM with advanced visual understanding and conversational capabilities. Comprehensive experiments and case studies demonstrate that OphIn-VL achieves superior performance compared with state-of-the-art general medical and domain-specific MLLMs.
\end{abstract}

\section{Introduction}
Recent advances of artificial intelligence (AI) have reshaped the trajectory of medical AI research~\cite{dong2025cunsb,dong2025tpot,khan2024recent,ali2026systematic,chen2025cracking,zhu2025toward, li2025evit, dong2025talk}. Specifically, single-objective tasks, such as image segmentation or disease classification, are no longer the sole focus of the research community. Instead, increasing attention has shifted toward developing systems that can address multiple tasks comprehensively and integrate into clinical workflows~\cite{moor2023foundation}. In this context, medical Multimodal Large Language Models (MLLMs) have emerged as a promising paradigm and have already demonstrated encouraging progress in clinical applications~\cite{dong2025llada,wang2022medclip,zhang2022contrastive}. Despite these advances, adapting MLLMs to specialized medical domains, such as ophthalmology, remains largely underexplored. Although ophthalmology has witnessed substantial progress in visual foundation models, most existing efforts still focus on learning image representations for downstream tasks~\cite{silva2025foundation,shi2024eyeclip,du2024ret,wu2024mm}. For example, RETFound~\cite{RETFound} adopts masked autoencoder (MAE)~\cite{he2022masked} pretraining on millions of retinal images and demonstrates strong performance in downstream applications such as disease detection. However, these foundation models are typically designed for representation learning rather than instruction following or interactive communication, limiting their direct applicability to clinical deployment. To address this limitation, recent studies have begun to explore the generative capabilities of ophthalmic AI through instruction tuning and post-training techniques.

Despite recent efforts to advance instruction-tuned ophthalmic MLLMs~\cite{gao2023ophglm, chen2024ffa, deng2026fundus}, two major bottlenecks remain in the field. First, existing ophthalmic instruction datasets are often synthesized from established public benchmarks rather than newly collected clinical sources. For example, MM-Retinal-Reason~\cite{wu2025bridging} improves ophthalmic reasoning by leveraging public datasets such as LAG~\cite{li2019attention}. While effective, this dependence on pre-existing benchmarks limits image-source novelty and raises concerns about clinical complexity and real-world applicability. Moreover, because these benchmarks were typically designed for predefined tasks (e.g., lesion segmentation), they often provide only task-specific annotations while lacking rich clinical information. Second, the restricted data sources further constrain dataset scale and dialogue diversity, which are essential for training robust conversational models. For instance, RetinalGPT~\cite{zhu2025retinalgpt} fine-tunes its language backbone using only 38K instruction conversations, resulting in relatively limited interaction formats and constrained clinical communication ability. Together, these limitations motivate two research questions: 
\begin{itemize}
    \item \textit{How can we curate large-scale ophthalmology-specific instruction data from novel clinical sources beyond existing public benchmarks?}
    \item \textit{Can such scalable and diverse instruction data improve the clinical reasoning and generalization capabilities of ophthalmic MLLMs?}
\end{itemize}
To address these challenges, we introduce OphIn-Engine, a novel ophthalmology-specific data curation pipeline for automatically generating instructional image-text pairs from real-world ophthalmic videos. The engine consists of four modules. First, the multimodal transcription module extracts raw image-transcript pairs from open-source web videos through keyframe selection, audio transcription, image refinement, and de-identification. Second, the visual cue separation and scoring module identifies clinically relevant visual features and filters low-quality instances. Third, the instruction synthesis module transforms the refined instances into visual question answering (VQA), multi-turn conversational, and chain-of-thought (CoT) instruction data. Finally, the post-verification module serves as quality control to remove ill-posed, inconsistent, or hallucinated generations. Using this scalable pipeline and broad web-based resources, we construct OphIn-500K, a large-scale ophthalmology instruction dataset containing 536,132 instances and 151,430 unique images, covering over 1,000 retinal pathology conditions and 800 ophthalmic anatomical features. To further evaluate the utility of this dataset, we fine-tune a vision-language backbone to develop OphIn-VL, an ophthalmology-specific MLLM. Comprehensive experiments and case studies demonstrate that OphIn-VL achieves strong ophthalmic visual interpretation performance and supports clinical multi-turn conversations.

In summary, our contributions are three-fold: (1): We introduce OphIn-Engine, an automated curation pipeline that transforms large-scale web-based ophthalmology videos into high-quality multimodal instruction pairs, reducing reliance on static public benchmarks. (2): Using OphIn-Engine, we construct OphIn-500K, a large-scale ophthalmology instruction dataset comprising over half a million instances that capture the complexity of authentic clinical scenarios and provide a robust resource for future ophthalmic research. (3): We develop OphIn-VL, a specialized MLLM trained on OphIn-500K, establishing a new direction for future research. Through systematic evaluations and case studies, we demonstrate its strong performance in both ophthalmic visual understanding and clinical conversational ability.

\section{Related Work}
\subsection{Medical Multimodal Datasets}
The success of general-purpose MLLMs in integrating visual perception with linguistic reasoning has inspired the medical research community to build specialized clinical assistants, leading to a surge in medical MLLMs~\cite{lin2025taming,moor2023med,li2023llava}. Driven by this momentum, the underlying dataset has undergone a significant paradigm shift. Specifically, single-task objectives (e.g., disease classification or anatomical segmentation) are no longer the sole focus of the field. Instead, recent research aims to advance the capabilities of medical AI by proposing complex Visual Question Answering (VQA) instances~\cite{zhang2023pmc,liu2021slake}, multi-turn conversation data~\cite{wang2023chatcad}, and high-quality Chain-of-Thought (CoT) datasets~\cite{le2025s,wang2025v2t}. For example, LLaVA-Med~\cite{li2023llava} extends training datasets into multi-turn conversational scenarios, enabling medical AI to interact naturally with users in diagnostic settings. Despite these successes, a critical gap continues to limit the development of ophthalmic AI. Most medical multimodal datasets are heavily dominated by classical radiology and pathology, leaving highly specialized domains like ophthalmology largely underexplored by general benchmarks~\cite{lozano2025biomedica,sevgi2025foundation}. Even when ophthalmic datasets do exist, their scale and image sources remain constrained. To bridge this gap and establish an automated data curation pipeline for ophthalmic research, we propose the OphIn-Engine. This pipeline directly leverages dynamic web resources to generate high-quality, authentic ophthalmic instruction data. Building on this engine, we introduce OphIn-500K, a large-scale instruction-tuning dataset designed to overcome existing data bottlenecks and significantly enrich the study of ophthalmic AI.

\subsection{Multimodal Large Language Models in Ophthalmology}
The advancement of general medical AI has catalyzed a revolution in ophthalmic research~\cite{van2025foundation,gurnani6artificial}. Specifically, the research community anticipates developing systems that not only assist in diagnostic tasks but also integrate into clinical workflows as interactive conversational agents~\cite{wang2025llm,holland2025specialized,sathya2025detection}. This trend is reflected in recent works that pivot toward developing specialized MLLMs capable of clinical interpretation and visual understanding. For example, FLAIR~\cite{silva2025foundation} moves beyond simple classification to achieve robust retinal image understanding through large-scale contrastive language-image pretraining. However, two bottlenecks persist in this domain. First, despite demonstrating strong visual understanding, many ophthalmology MLLMs remain heavily focused on feature learning, lacking the capacity to robustly follow complex human instructions, generate reports, or engage in clinical conversations~\cite{RETFound,silva2025foundation,du2024ret,shi2024eyeclip,yu2024urfound}. Second, even when more recent models transition to generative architectures and undergo instruction tuning or post-training, they are often hindered by training data scale and image diversity, resulting in constrained clinical utility~\cite{holland2025specialized,hartsock2024vision}. For instance, while Fundus-R1~\cite{deng2026fundus} aims to improve reasoning abilities through verifiable rewards, it relies on approximately 168,000 training samples sourced exclusively from highly reused public benchmarks. Consequently, a gap in training data scale and image novelty persists when compared to the state-of-the-art general medical MLLMs. Given this context, we introduce OphIn-VL, our ophthalmologic-specific MLLM. Benefiting from the unprecedented scale and real-world complexity of the OphIn-500K training data, our model demonstrates great visual understanding and conversational capabilities.

\section{OphIn-Engine}\label{sec:dataset}
\begin{figure*}[t]
    \centering
    \includegraphics[width=0.9\linewidth]{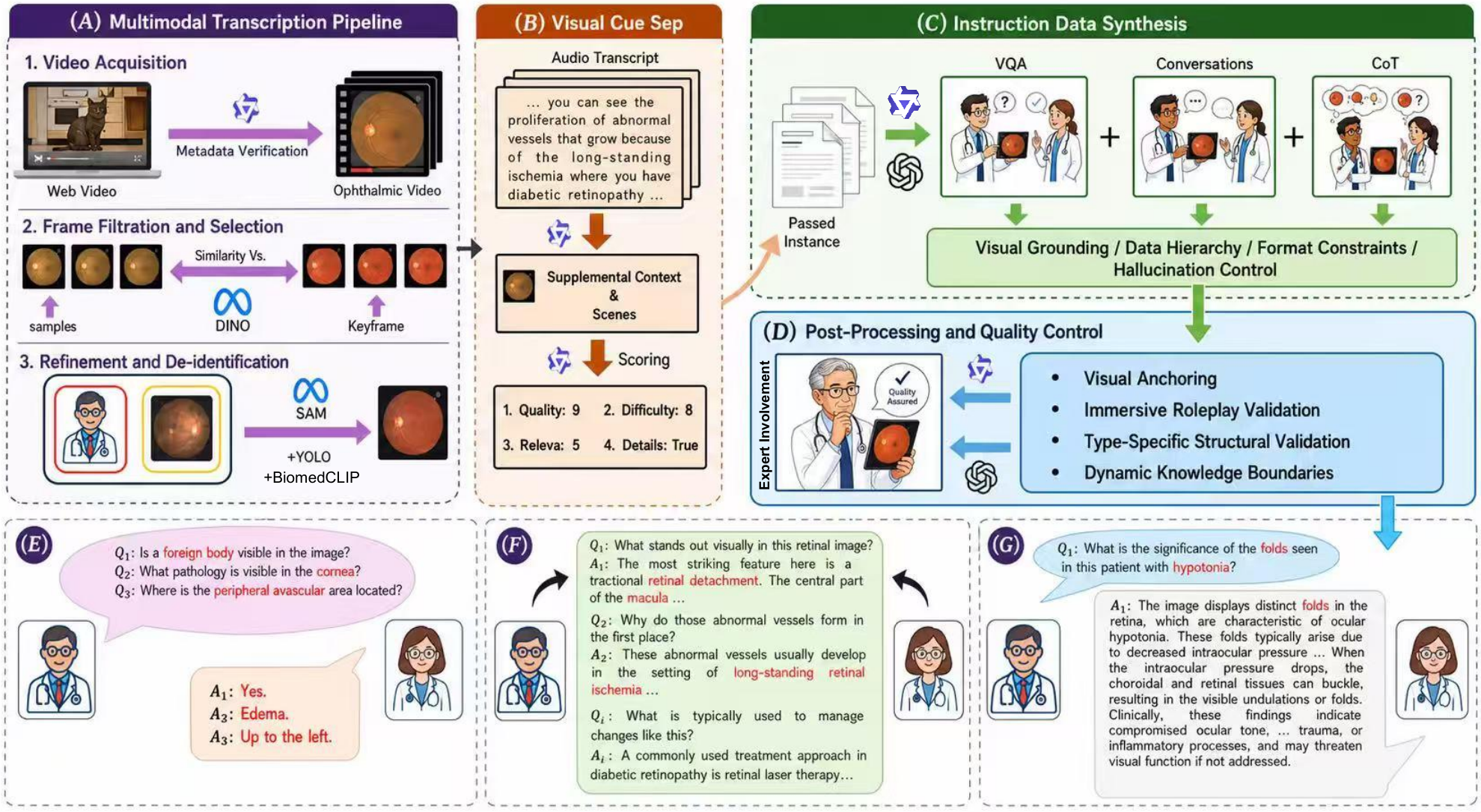}
    \caption{Overview of the \textbf{OphIn-Engine}. \textbf{(A)} illustrates the multimodal transcription pipeline. \textbf{(B)} details the visual cue separation and scoring procedure applied to the audio transcripts. \textbf{(C)} depicts the instruction data synthesis pipeline, which generates VQA, conversational interactions, and CoT data. \textbf{(D)} Outlines the final post-processing and quality control protocols (e.g., visual anchoring) used to filter out failure instances, followed by \textbf{expert-involved evaluation} for further quality assessment.\textbf{(E)}, \textbf{(F)}, and \textbf{(G)} present sample outputs of the retained VQA, multi-turn conversations, and CoT data, respectively. The images are provided solely for visualization purposes. }
    \label{fig:data-pipeline-whole}
\end{figure*}

This section details the architecture of the OphIn-Engine. As illustrated in Fig.~\ref{fig:data-pipeline-whole}, the engine comprises four core modules. First, the multimodal transcription pipeline (Sec.~\ref{subsec:sft-firststep}) collects and processes the raw video clips. Next, the visual cue separation and scoring component (Sec.~\ref{subsec: sft-preprocess}) is designed to filter and refine the noisy audio transcripts. The instruction data synthesis module (Sec.~\ref{subsec: sft-data-curation}) then executes our specific data curation methodology. Finally, the post-processing and quality control stage (Sec.~\ref{subsec:sft-quality-control}) serves as a robust safeguard against hallucinations and ill-posed generations. Due to space constraints, detailed implementation specifics are provided in \textbf{Appendix A}.

\subsection{Multimodal Transcription Pipeline}\label{subsec:sft-firststep}
\noindent \textbf{Video acquisition, clinical frame filtering, and audio transcription.} We retrieve candidate videos via the Open-access Data using a custom keyword dictionary. This ensures the data originates from professional ophthalmic contexts (e.g., symposia) and strictly focuses on three imaging modalities: Color Fundus Photography (CFP), Optical Coherence Tomography (OCT), and Ultra-Widefield Fundus (UWF). The retrieved metadata is then evaluated by an LLM to guarantee clinical relevance and informational density. To extract the target images, we employ a two-stage frame filtering process. First, we uniformly sample frames from the videos and extract their embeddings using a frozen DINO~\cite{caron2021emerging} backbone. Candidate video episodes are segmented when the cosine similarity between consecutive frame embeddings falls below a predefined threshold. As shown in Fig.~\ref{fig:data-pipeline-whole}(A), these sharp fluctuations in the embedding space serve as boundary indicators for visual content shifts. By detecting these transitions (e.g., moving from a detailed CFP analysis to a general panel discussion), we segment the video into highly cohesive, topic-specific episodes, ensuring that the paired audio transcript remains contextually pure. Next, the frames within each episode are processed by a fine-tuned DINO classifier to detect the presence of retinal imaging. Given the high semantic consistency within each episode, we select the frame with the highest probability of containing medical imaging to represent the entire segment, forming our final image corpus. Correspondingly, we use Whisper~\cite{whisper2022} to generate text transcripts from the raw audio associated with each identified episode. To prevent abrupt cutoffs and ensure the transcript captures the complete clinical context, we incorporate an additional temporal buffer at the episode boundaries.

\noindent \textbf{Image Refinement and De-identification.} Due to the inherently noisy nature of web-sourced imagery (e.g., keyframes containing both retinal scans and decorative graphics), we conduct an additional refinement step to ensure visual consistency. As shown in Fig.~\ref{fig:data-pipeline-whole}(A), we first employ the SAM~\cite{kirillov2023segment,ma2024segment} to extract candidate sub-regions from each keyframe, defaulting to the entire frame if no distinct regions are detected. We then extract features for these sub-regions using BiomedCLIP~\cite{Biomedclip} and compare them against predefined text dictionaries. A sub-region is retained only if its similarity score exceeds the threshold for at least one positive clinical prompt (e.g., "diabetic retinopathy") and remains below the threshold for all negative prompts (e.g., "portrait photograph"). All discarded sub-regions are subsequently masked out, and the image is cropped accordingly. Finally, to ensure strict privacy compliance, we utilize a YOLO-based~\cite{YOLOv8} detector to localize any remaining sensitive information (e.g., patient faces). Detected sensitive areas are completely masked while carefully preserving the underlying clinical features.

\subsection{Visual Cue Separation and Preprocessing}\label{subsec: sft-preprocess}
\begin{figure}[h]
    \centering
    \includegraphics[width=0.9\linewidth]{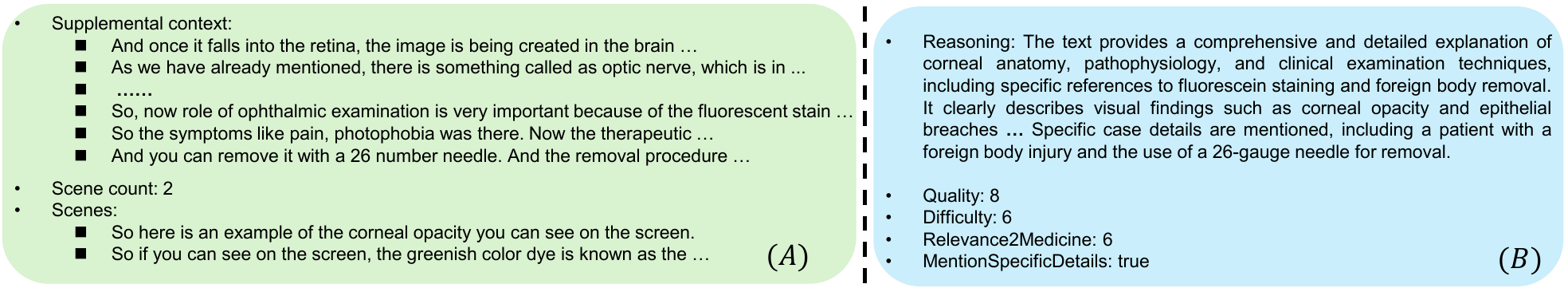}
    \caption{Illustration of the raw transcript preprocessing. \textbf{(A)} represents a result of the visual cue separation. \textbf{(B)} denotes an output of the quality scoring process. }
    \label{fig:pipeline-preprocessing}
\end{figure}

\noindent \textbf{Visual cue separation.} The data acquisition pipeline detailed in Sec.~\ref{subsec:sft-firststep} yields raw, unrefined image-text pairs. Since these raw audio transcripts frequently include conversational fillers and general medical background that do not directly correspond to the visual content, we prompt an LLM judge to decouple the raw text into two distinct components: \texttt{Scenes} and \texttt{Supplemental Context}. This step ensures that the generated data is grounded in visual features rather than hallucinated from the language model's prior knowledge. Acting as a strict data extractor, the judge simply partitions the content without any modification or paraphrasing. As shown in Fig.~\ref{fig:pipeline-preprocessing}(A), \texttt{Scenes} strictly isolate the explicit visual cues and descriptions present in the image (e.g., observable corneal opacity). Conversely, the \texttt{Supplemental Context} retains the broader clinical discourse and non-visual metadata (e.g., patient symptoms such as photophobia or related procedures).

\noindent \textbf{Quality scoring.} Following visual cue separation, we prompt a second LLM judge to evaluate the quality of each transcript. Guided by clinical experts, the model assesses the text across four metrics and generates supporting rationales: \texttt{Quality}, \texttt{Difficulty}, \texttt{Relevance2Medicine}, and the boolean \texttt{MentionSpecificDetails}. Specifically, \texttt{Quality} evaluates the clarity and descriptive power of the text. \texttt{Difficulty} assesses the clinical complexity. \texttt{Relevance2Medicine} measures the clinical relevance of the text to ophthalmology. Finally, \texttt{MentionSpecificDetails} verifies again whether the text contains specific case details or anatomical landmarks that serve as clear visual cues. For example, as shown in Fig.~\ref{fig:pipeline-preprocessing}(B), a high-quality instance explicitly mentions visual markers like corneal opacity, alongside complementary clinical context, such as corneal anatomy. Ultimately, we retain only ophthalmology-relevant data containing explicit visual cues to construct the final instruction dataset.

\subsection{Instruction Synthesis}\label{subsec: sft-data-curation}
Building upon the scored data instances and their cleanly separated visual content, the OphIn-Engine categorizes the synthesized instruction data into three distinct formats: Visual Question Answering (VQA), multi-turn conversational interactions, and Chain-of-Thought (CoT)~\cite{wei2022chain} reasoning.

\noindent \textbf{Simple VQA.} This subset is designed to refine the model's ophthalmology-specific comprehension, ensuring responses are explicitly grounded in visual features rather than hallucinated from prior knowledge. We employ an LLM generator, prompted to formulate question-answer pairs based exclusively on the visual cues within the \texttt{Scenes} component. To ensure objective grounding, it must construct auxiliary reasoning chains to justify its outputs and is strictly prohibited from referencing the raw transcripts. To maximize linguistic diversity, we mandate the generation of three question types per instance: \texttt{Yes/No}, \texttt{What} and \texttt{Where}. Furthermore, to mitigate hallucination, the model is instructed to output \texttt{N/A} if the visual metadata is insufficient to support a factual QA pair. As illustrated in Fig.~\ref{fig:data-pipeline-whole}(E), concise answers (e.g., up to the left) are derived directly from factual evidence within the \texttt{Scenes} (e.g., "we notice a peripheral avascular area up to the left"). Throughout this process, the \texttt{Supplemental Context} serves solely as a passive medical reference to clarify ambiguous clinical terms.

\noindent \textbf{Conversational interactions.} To enhance the model's capacity for expert-level clinical dialogue, we introduce an additional LLM generator to synthesize a multi-turn conversational subset. Specifically, the generator is prompted to construct each dialogue as if the participants are actively examining the provided image. To guarantee strict visual grounding, the opening turn must be formulated using the visual cues from the \texttt{Scenes} metadata exclusively. Subsequently, unlike in a standard VQA dataset, the later turns may incorporate information from the \texttt{Supplemental Context} to facilitate deeper clinical discourse on disease mechanisms, staging, or treatment options. As illustrated in Fig.~\ref{fig:data-pipeline-whole}(F), the initial turns focus strictly on observable visual findings (e.g., retinal detachment or long-standing retinal ischemia). The dialogue then naturally evolves into complex clinical management inquiries (e.g., "retinal laser therapy").

\noindent \textbf{CoT reasoning.} Diagnostic reasoning requires explicit logical progression. Consequently, we prompt an LLM generator to synthesize single-turn CoT data using only our highest-quality instances. For this subset, it formulates questions based on visual findings and structures the responses around a three-phase logical chain: (1) the observation phase, which details specific visual findings; (2) the correlation phase, which connects these findings to underlying clinical mechanisms; and (3) the conclusion phase, which provides the answer alongside a clinical explanation. Importantly, we permit the generator to leverage its internal medical knowledge to bridge these logical transitions once the transcript lacks sufficient depth. As shown in Fig.~\ref{fig:data-pipeline-whole}(G), the query is grounded in specific visual findings (e.g., folds and hypotonia). The corresponding response naturally acknowledges the folds, discusses potential clinical causes, and concludes with an assessment of ocular tone.

\subsection{Post-Processing and Quality Control}\label{subsec:sft-quality-control}
\begin{figure}[ht]
    \centering
    \includegraphics[width=0.9\linewidth]{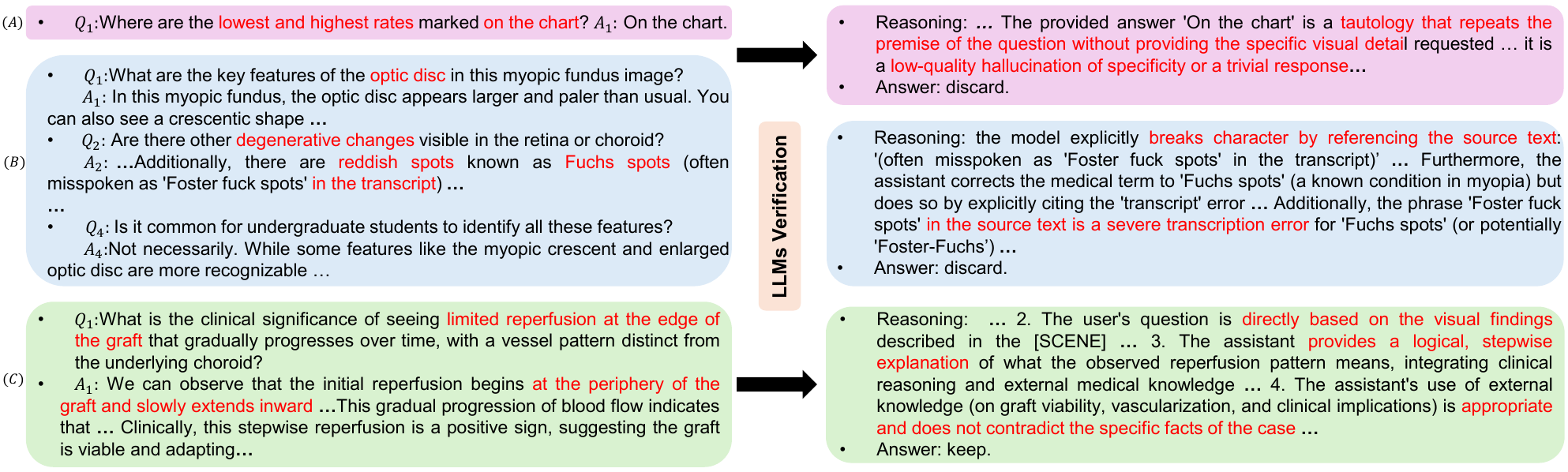}
    \caption{Illustration of the final data post-processing and quality control stages. \textbf{(A)} shows verification for \texttt{Where}-type VQA samples. \textbf{(B)} presents the verification results for conversational interaction data. \textbf{(C)} shows the evaluation results for CoT data. }
    \label{fig:postprocessing-exp}
\end{figure}
Following data synthesis, the engine implements a final LLM-based verification stage to filter out failure generation, including ill-posed questions, hallucinated content, and persona-breaking responses (See Fig.~\ref{fig:data-pipeline-whole}(D)). For the simple VQA set, we first discard unanswerable queries (i.e., those outputted as \texttt{N/A}). We then employ an LLM verifier to confirm that each QA pair is strictly grounded in visual evidence, devoid of transcript reliance or fabricated details. As illustrated in Fig.~\ref{fig:postprocessing-exp}(A), $F$ rejects responses lacking visual specificity (e.g., “on the chart”) for failing this grounding criterion. For conversational data, beyond enforcing visual grounding in the opening turn and preserving character consistency throughout, we require dialogues to avoid hallucinating case-specific facts, such as fictitious patient histories or unobservable visual findings. Conversely, we permit the integration of general medical knowledge (e.g., common side effects) provided it enriches the interaction without contradicting the visual case. As shown in Fig.~\ref{fig:postprocessing-exp}(B), conversations are discarded if they break persona by explicitly referencing the source text (e.g., mentioning "Foster spots" as if reading from notes). Finally, for CoT reasoning, we apply similar grounding criteria but additionally discard instances that lack intermediate logical steps or exhibit overly rigid, templated formatting. Ultimately, we retain only instances explicitly labeled with a \texttt{keep} decision by the verifier (see Fig.~\ref{fig:postprocessing-exp}(C)). After this automated filtering stage, experts are further involved through sample-based verification to provide an additional assessment.

\section{OphIn-500K}\label{subsec:data-statistic}

\begin{table}[htbp]
\centering

\resizebox{0.9\columnwidth}{!}{ 
\begin{tabular}{@{} l l r | r c c c @{}}
\toprule
\multirow{2}{*}{\textbf{Data type}} & \multirow{2}{*}{\textbf{Subtype}} & \multirow{2}{*}{\textbf{Instances}} & \multirow{2}{*}{\textbf{Images}} & \multicolumn{3}{c}{\textbf{Modality Number}} \\
\cmidrule(lr){5-7} 
& & & & CFP & OCT & UWF \\
\midrule

\multirow{3}{*}{VQA} 
& Yes/No & 145,252 & 145,252 & 37,738 & 76,207 & 31,307 \\
& What   & 142,971 & 142,971 & 37,202 & 74,914 & 30,855 \\
& Where  & 110,898 & 110,898 & 28,817 & 58,221 & 23,860 \\
\midrule

Conversation & \textit{N/A} & 124,441 & 124,441 & 32,006 & 65,412 & 27,023 \\

CoT          & \textit{N/A} & 12,570  & 12,570  & 2,745 & 6,883 & 2,942 \\
\midrule

\textbf{Total} & \textit{N/A} & \textbf{536,132} & \textbf{151,430} & \textbf{38,943} & \textbf{80,139} & \textbf{32,348} \\

\bottomrule
\end{tabular}
}
\caption{Statistical overview of the OphIn-500K dataset. This breakdown details the instance counts across all categories. For each data type, we specify the total image count alongside the distribution of modalities, including CFP, OCT, and UWF imaging.}
\label{tab:sft-datastatis}
\end{table}
Processed through the OphIn-Engine, the curation pipeline yields over 500,000 high-quality instances extracted from approximately 29,465 video clips, spanning roughly 14,000 hours of footage. As detailed in Tab.~\ref{tab:sft-datastatis}, the OphIn-500K dataset comprises 536,132 instances derived from 151,430 unique images. This dataset encompasses three primary ophthalmological modalities essential for retinal diagnosis: CFP, OCT, and UWF. Leveraging diverse web resources, it covers over 1,000 retinal conditions (e.g., macular edema, cataracts) and approximately 800 distinct anatomical features (e.g., crystalline lens, ciliary body), sourced from clinical cases across 100 different countries and regions. Notably, unlike existing ophthalmology-specific datasets, OphIn-500K introduces large-scale image-text pairs without relying on heavily recycled public benchmarks. To rigorously evaluate model performance, we introduce the OphIn-VQA evaluation split. Specifically, we sample VQA instances from the dataset while ensuring comprehensive coverage of major diseases and anatomical features. This yields 1,234 testing instances, consisting of 450 \texttt{Yes/No}, 451 \texttt{What}, and 333 \texttt{Where} open-ended questions, with the remaining VQA data retained exclusively for training. Ultimately, by bypassing static public datasets in favor of extracting tens of thousands of real-world frames, this dataset achieves a scale and visual novelty previously unseen in ophthalmic AI. Due to space constraints, a comprehensive statistical analysis is provided in \textbf{Appendix B}.

\section{OphIn-VL}
Building upon OphIn-500K, we introduce OphIn-VL, a novel, ophthalmology-specific MLLM that demonstrates robust visual understanding and the capacity for nuanced clinical dialogue. This section details our experimental setup (Sec.~\ref{subsec:exp-setup}), analyzes the primary performance results (Sec.~\ref{subsec:exp-analysis}), and provides further in-depth evaluations (Sec.~\ref{subsec:further-analysis}). Due to space constraints, comprehensive implementation details are deferred to \textbf{Appendix C}.

\subsection{Experimental Setup}\label{subsec:exp-setup}

\noindent \textbf{Training configurations.} OphIn-VL selects Qwen-3.5-9B~\cite{qwen3.5} as its core language backbone, adopting the SWIFT~\cite{zhao2024swiftascalablelightweightinfrastructure} training pipeline for its consistency and efficiency. During training, the vision encoder remains completely frozen. To minimize computational overhead, we apply Low-Rank Adaptation (LoRA)~\cite{hu2022lora} to all linear modules in both the vision-language projector and the language backbone, using a rank of $r=32$ and a scaling factor of $\alpha=64$. The model is fine-tuned for 2 epochs utilizing the AdamW optimizer, an initial learning rate of $2\times 10^{-4}$, and a cosine learning rate schedule. We establish a maximum sequence length of 4096 tokens and a per-device batch size of 4. Training is accelerated using DeepSpeed ZeRO-2~\cite{rajbhandari2020zero} distributed across $4\times$NVIDIA A100 (80GB) GPUs. The complete training dataset comprises 534,898 unique instances, encompassing general conversational interactions and CoT reasoning data, alongside categorized VQA samples: 144,802 \texttt{Yes/No}, 142,520 \texttt{What}, and 110,565 \texttt{Where} instances. 

\noindent \textbf{Evaluation configurations and baselines.} As detailed in Sec.~\ref{subsec:data-statistic}, we evaluate model performance on the OphIn-VQA split. Because all three question categories are formulated as open-ended VQA, we employ an LLM-as-a-judge framework to assess answer accuracy. Additionally, we report the F1 score of BERTScore~\cite{zhang2019bertscore} to evaluate the semantic similarity. To mitigate language evaluator biases (e.g., a preference for verbosity) and handle variations in baseline output styles (e.g., concise formatted answers versus lengthy conversational responses), an auxiliary LLM extractor first distills all model responses into concise answers without altering their factual content. To ensure a robust evaluation, we compare our model against a diverse set of open-source baselines. General medical MLLMs include LLaVA-Med~\cite{li2023llava}, MedGemma~\cite{sellergren2025medgemma}, Lingshu~\cite{xu2025lingshu}, and MedVLM-R1~\cite{pan2025medvlm}. Ophthalmology-specific MLLMs include RetinalGPT~\cite{zhu2025retinalgpt}, FundusExperts~\cite{liu2025constructing}, and OphthaReason~\cite{wu2025bridging} series. Finally, we evaluate general-domain MLLMs, including the Qwen~\cite{qwen3.5}, LLaMA~\cite{grattafiori2024llama}, and InternVL~\cite{chen2024internvl}, to establish a comprehensive performance baseline.

\begin{figure}[ht]
    \centering
    \includegraphics[width=0.9\linewidth]{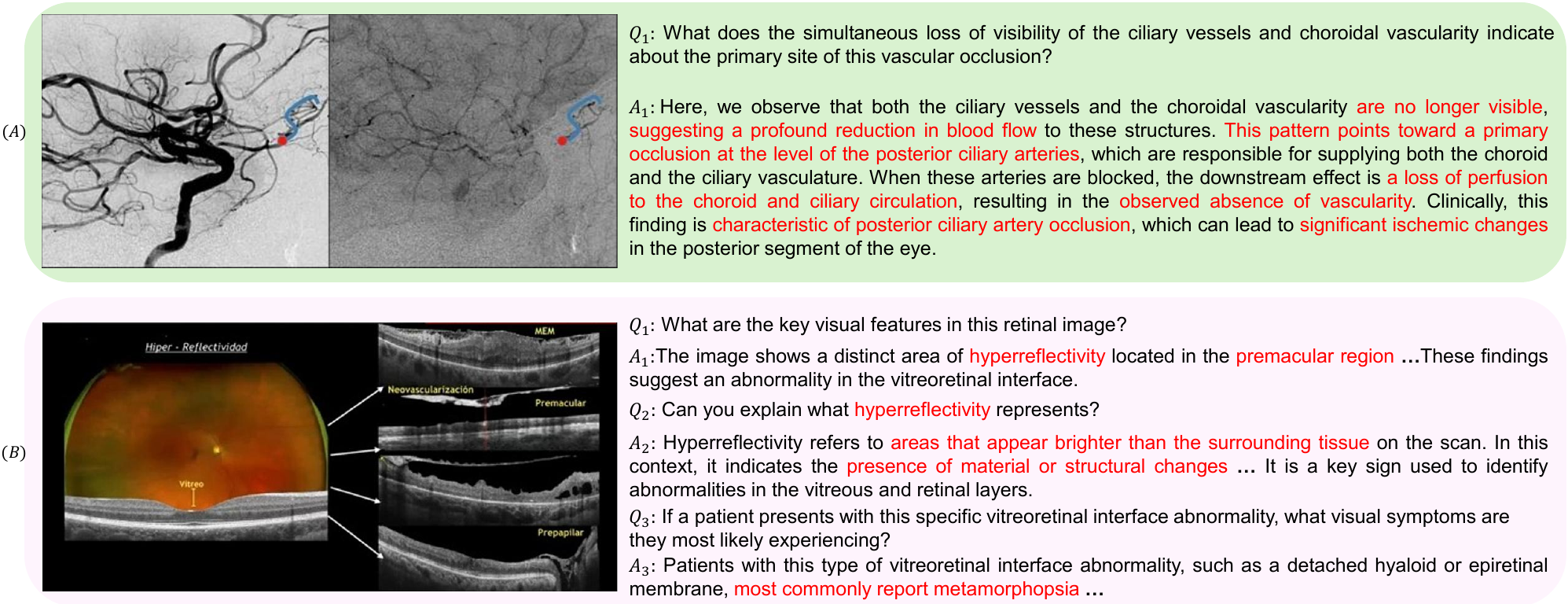}
    \caption{Illustration of OphIn-VL responses for retinal visual understanding. \textbf{(A)} shows the model response in a single-turn interaction, while \textbf{(B)} presents its responses in a multi-turn conversational setting. The red highlights the important information. The image–text queries are constructed from the OphIn-VQA split.}
    \label{fig:ours-example}
\end{figure}

\subsection{Experiment Results}\label{subsec:exp-analysis}
\begin{table*}[htbp] 
    \centering
    \renewcommand{\arraystretch}{1.2}
    \resizebox{0.9\textwidth}{!}{%
        \begin{tabular}{@{}ll cccccccc@{}}
            \toprule
            \multirow{2}{*}{\textbf{Category}} & \multirow{2}{*}{\textbf{Model}} & \multicolumn{2}{c}{\textbf{Yes/No}} & \multicolumn{2}{c}{\textbf{What}} & \multicolumn{2}{c}{\textbf{Where}} & \multicolumn{2}{c}{\textbf{Average}} \\
            \cmidrule(lr){3-4} \cmidrule(lr){5-6} \cmidrule(lr){7-8} \cmidrule(lr){9-10}
            & & \textbf{LLM} & \textbf{B-score} & \textbf{LLM} & \textbf{B-score} & \textbf{LLM} & \textbf{B-score} & \textbf{LLM} & \textbf{B-score} \\
            \midrule
            
            \multirow{5}{*}{General Medical MLLMs} 
            & LLaVAMed~\cite{li2023llava}     & 95.33 & \textbf{99.51} & 25.44 & 74.29 & 29.50 & 76.99 & 52.03 & 84.22 \\
            & MedGemma~\cite{sellergren2025medgemma}     & 48.67 & 94.66 & 27.11 & 75.64 & 34.76 & 75.59 & 37.03 & 82.57 \\
            & Lingshu~\cite{xu2025lingshu}       & 68.44 & 96.52 & 34.75 & 77.18 & 39.34 & 76.67 & 48.28 & 84.09 \\
            & MedVLM-R1~\cite{pan2025medvlm}     & 17.11 & 90.79 & 25.38 & 70.97 & 32.43 & 75.74 & 24.27 & 79.48 \\
            \midrule
            
            \multirow{3}{*}{Domain-specific MLLMs} 
            & RetinalGPT~\cite{zhu2025retinalgpt}    & 61.44 & 95.67 & 10.64 & 73.03 & 25.15 & 73.67 & 33.08 & 81.36 \\
            & OphthaReason-Qwen~\cite{wu2025bridging}  & 73.11 & 97.20 & 25.39 & 76.48 & 34.16 & 73.14 & 45.16 & 83.13 \\
            & OphthaReason-InternVL~\cite{wu2025bridging}  & 50.89 & 90.72 & 23.28 & 75.17 & 31.38 & 73.67 & 35.53 & 80.44 \\
            & FundusExperts~\cite{liu2025constructing} & 4.22 & 90.05 & 25.83 & 71.96 & 38.74 & 75.51 & 21.43 & 79.49 \\
            \midrule
            
            \multirow{3}{*}{General MLLMs} 
            & Qwen3.5-9B~\cite{qwen3.5}    & 48.78 & 71.72 & 35.86 & 69.60 & 49.32 & 70.68 & 44.21 & 70.66 \\
            & LLaMA~\cite{grattafiori2024llama}         & 59.67 & 91.13 & 26.22 & 74.10 & 34.08 & 67.74 & 40.54 & 78.59 \\
            & InternVL~\cite{chen2024internvl} & 70.28 & 87.94 & 34.76 & 75.00 & 39.34 & 73.01 & 49.25 & 79.17 \\

            \midrule
            \cellcolor[gray]{0.85}Ours & \cellcolor[gray]{0.85} OphIn-VL & \cellcolor[gray]{0.85} \textbf{97.33} & \cellcolor[gray]{0.85} 90.04 & \cellcolor[gray]{0.85} \textbf{54.71} & \cellcolor[gray]{0.85}\textbf{82.22} & \cellcolor[gray]{0.85} \textbf{53.75} & \cellcolor[gray]{0.85}\textbf{79.61} & \cellcolor[gray]{0.85}\textbf{70.00} & \cellcolor[gray]{0.85}\textbf{84.37}  \\
            
            \bottomrule
        \end{tabular}%
    }
        \caption{Performance on the OphIn-VQA evaluation set. The dataset comprises \texttt{Yes/No}, \texttt{What}, and \texttt{Where} question types across CFP, OCT, and UWF imaging modalities. Evaluation metrics include the LLM-as-a-judge score (i.e., LLM) and F1 score of BERTScore (i.e., B-score). Best results are highlighted in \textbf{bold}.}
    \label{tab:vqa_model_comparison}
\end{table*}
The strong performance of OphIn-VL demonstrates the effectiveness of OphIn-500K. As shown in Tab.~\ref{tab:vqa_model_comparison}, OphIn-VL outperforms competing models across all three question types, indicating strong ophthalmic visual understanding capability. Specifically, OphIn-VL achieves 97.33\% on the \texttt{Yes/No} evaluation split, outperforming both general medical MLLMs and ophthalmology-specific MLLMs by a clear margin. When the evaluation shifts to more challenging splits that require advanced spatial reasoning, all models exhibit noticeable performance degradation. Nevertheless, OphIn-VL maintains a substantial advantage. For example, MedGemma achieves only 34.76\% on the \texttt{Where} split, whereas OphIn-VL improves the score to 53.75\%. Additionally, when the image sources extend beyond well-established public benchmarks, on which many existing ophthalmology-specific MLLMs are built, the performance gap between domain-specific MLLMs and general medical MLLMs becomes less pronounced. For instance, OphthaReason-Qwen achieves an average score of 45.16\%, while Lingshu reaches 48.28\%. This finding suggests that current ophthalmology-specific MLLMs may still be limited by insufficient data diversity and may not generalize well to novel image sources.

We further provide case studies to demonstrate the ability of OphIn-VL to support clinically grounded multi-turn interactions. As shown in Fig.~\ref{fig:ours-example}(A), in a single-turn dialogue requiring complex reasoning, such as localization-based interpretation, OphIn-VL naturally progresses from visual observation (e.g., "no longer visible") to its association with relevant clinical structures (e.g., posterior ciliary arteries), and ultimately reaches an accurate diagnostic implication (e.g., ischemic changes of the eye). In addition, OphIn-VL also demonstrates the ability to engage in retinal-domain conversations. As shown in Fig.~\ref{fig:ours-example}(B), the model first identifies hyperreflectivity in the premacular region, then explains its clinical definition (e.g., brighter area) and further discusses its ophthalmological significance (e.g., indicating the presence of structural changes).

\subsection{Further Discussions}\label{subsec:further-analysis}

\begin{figure}[t]
    \centering
    \includegraphics[width=0.95\linewidth]{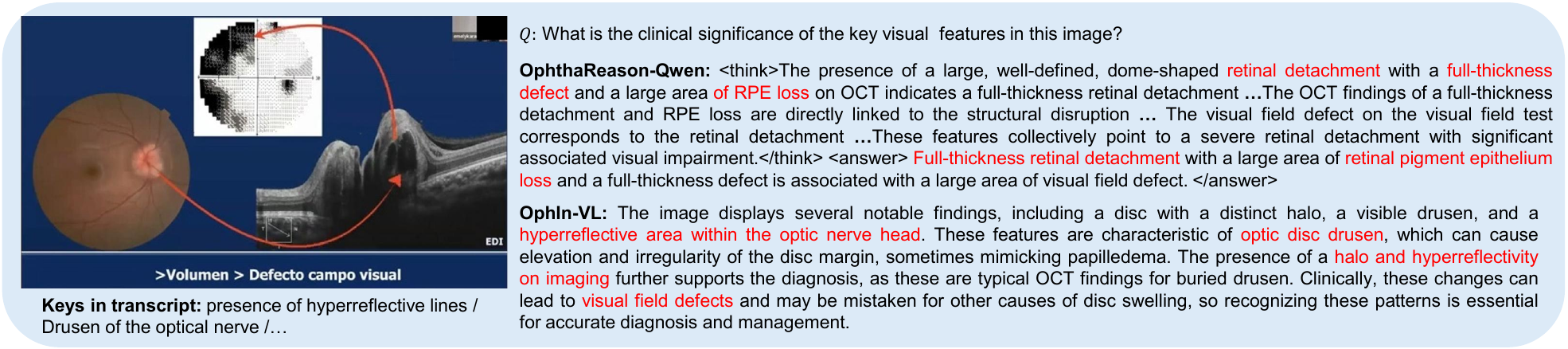}
    \caption{A comparison of conversational performance between OphIn-VL and OphthaReason-Qwen in a single-turn setting. Key clinical information is highlighted in red. The representative image-query pair was synthesized from the OphIn-VQA split.}
    \label{fig:final-compare}
\end{figure}

\begin{table}[ht]
\centering
\resizebox{0.9\columnwidth}{!}{
\begin{tabular}{lccc}
\toprule
\textbf{Dataset}  & \textbf{Scale} & \textbf{Question Form} & \textbf{Image Source} \\
\midrule
MM-Retinal-Reason~\cite{wu2025bridging}  & 300K & VQA / Caption / CoT & Public benchmarks \\
FundusGen~\cite{liu2025constructing}  & 300K & VQA / Conv / CoT & Public benchmarks \\
RetinalGPT~\cite{zhu2025retinalgpt}  & 76K & VQA / Conv & Public benchmarks \\
\midrule
\textbf{OphIn-500K}  & \textbf{500K} & 
\textbf{VQA / Conv / CoT} & 
\textbf{Newly collected} \\
\bottomrule
\end{tabular}
}
\caption{Comparison of OphIn-500K with existing domain-specific instruction datasets. OphIn-500K is novel in both scale and image sources.}
\label{tab:dataset_comparison}
\end{table}

\noindent \textbf{The uniqueness of OphIn-500K.} Rooted in OphIn-Engine (Sec.~\ref{sec:dataset}), OphIn-500K is distinctive in both data source and scale. As shown in Tab.~\ref{tab:dataset_comparison}, many existing ophthalmology-specific instruction datasets (e.g., FundusGen) are synthesized from established public benchmarks and are therefore constrained by the scale and diversity of the underlying images. In contrast, by leveraging abundant web-based resources, OphIn-500K overcomes the limitation of repeatedly reusing existing benchmarks and enables to be a large-scale instruction dataset for advancing ophthalmology AI research.

\noindent \textbf{Case comparison.} Typical domain-specific MLLM face challenges in more complex scenarios. As shown in Fig.~\ref{fig:final-compare}, while OphthaReason-Qwen demonstrates a rational reasoning chain but reaches hallucinated conclusions (e.g., retinal detachment and RPE loss), OphIn-VL, on the other hand, correctly recognizes the key factors (e.g., optic disc drusen), leading to final conclusions (e.g., visual field defects). 

\section{Conclusion}
In this work, we presented OphIn-Engine, an automated and scalable data curation pipeline designed to generate high-quality multimodal instruction pairs from real-world ophthalmic web videos. Using this pipeline, we constructed OphIn-500K, a large-scale ophthalmology instruction dataset comprising over half a million instances and covering more than 1,000 retinal conditions, offering a scalable alternative to instruction data derived from recycled public benchmarks. To demonstrate the utility of OphIn-500K, we further developed OphIn-VL, a domain-specific MLLM for ophthalmology. Comprehensive evaluations and case studies show that OphIn-VL achieves strong ophthalmic visual interpretation performance and strong multi-turn clinical conversational capabilities. We hope this work provides a valuable foundation for advancing next-generation ophthalmic AI systems.

\section*{Limitation} This work has several limitations. First, OphIn-500K currently focuses on CFP, OCT, and UWF, and future extensions could incorporate additional modalities to further improve data diversity and support broader ophthalmology AI research. Second, the current instruction data mainly consists of VQA, multi-turn conversations, and CoT samples. Future work could explore more rigorous instruction formats, including VQA paired with auxiliary reasoning chains for post-training alignment and preference data designed to improve clinical reliability.

\bibliography{custom}

\appendix

\section{Implementation Details of OphIn-Engine}
\subsection{Implementation of Multimodal Transcription Pipeline}
We retrieve candidate videos from open-access sources using a custom ophthalmology keyword dictionary. During retrieval, we collect metadata from three major aspects. First, we record semantic information, including video titles, descriptions, tags, and durations. Second, we perform quality and compliance checks, such as verifying the availability of audio captions and licensing information to support downstream processing and open-source compliance. Third, we assess source authority using channel-level information. These metadata help ensure that the collected videos originate from professional ophthalmic contexts, such as academic lectures and symposia, and focus on three target imaging modalities: color fundus photography (CFP), optical coherence tomography (OCT), and ultra-widefield fundus imaging (UWF). After constructing the initial metadata pool, we remove failure cases, such as corrupted videos or videos that are too short. We then use an LLM judge, Qwen-3-4B~\cite{qwen3}, to further evaluate the semantic metadata and filter out content unrelated to general medicine or ophthalmology.

After metadata-level preprocessing, we apply a two-stage frame filtering procedure. First, we uniformly sample frames from each video and encode them using a frozen DINO backbone~\cite{caron2021emerging}, specifically DINOv3~\cite{DINOv3}. We segment each video into candidate episodes by detecting sharp visual transitions, defined as cases where the cosine similarity between consecutive frame embeddings falls below a threshold of $\delta_c = 0.95$. This step divides long videos into coherent, topic-specific episodes. Next, to select representative keyframes with sufficient retinal information, we train a three-layer MLP classifier on top of the frozen DINOv3 encoder. The frame with the highest predicted retinal-imaging probability within each episode is selected as its keyframe. The classifier is trained using manually selected frames from the initial video frame pool. For audio processing over selected episodes, we use Faster-Whisper Large-v3~\cite{whisper2022} to generate raw transcripts. To avoid abrupt truncation and preserve complete clinical context, we add a temporal buffer of $\delta_t = 3$ seconds to both sides of each episode boundary before transcription.

After obtaining the initial keyframe-transcription pairs, we perform an additional image refinement step to improve visual consistency. Specifically, we first apply SAM~\cite{ma2024segment} (i.e., SAM3) to extract candidate sub-regions from each keyframe, using the full frame as a fallback when no distinct region is detected. We then encode these sub-regions with frozen BiomedCLIP~\cite{Biomedclip} and compare their features against predefined positive and negative text dictionaries. A sub-region is retained only if its similarity exceeds a threshold for at least one positive clinical prompt, such as diabetic retinopathy, while remaining below the threshold for all negative prompts, such as portrait photograph. Discarded regions are masked out, and the image is cropped around the retained clinical content. If a keyframe contains multiple retained sub-regions, we crop and incorporate all of them while preserving their relative spatial positions. Finally, to ensure privacy compliance, we use a YOLOv8 detector~\cite{YOLOv8} to localize potentially sensitive information, such as patient faces. Detected sensitive regions are fully masked while preserving the relevant clinical imaging content.

\begin{promptbox}[label=prompt:system-visualcue]{System prompt for visual cue separation.}
You are an expert ophthalmic data curator processing raw transcripts from medical video presentations. Your task is to take a raw paragraph of text and extract ONLY the explicitly visible clinical features and their spatial cues. You must also determine if the speaker is describing a single image or transitioning between multiple images on a slide, and split the data accordingly.

\textbf{STRICT FILTERING RULES:}

- 1. DO NOT PARAPHRASE. All extracted text must be verbatim substrings from the original transcript.

- 2. MERGE CUES AND FINDINGS: Do not split pointing words (e.g., "on the top") from the findings they describe (e.g., "neovascular lesion"). Keep them together as a single, natural phrase.

- 3. SAFE SUBSTRING EXTRACTION FOR MIXED SENTENCES: If a single sentence mixes a visible clinical fact with a future prognosis or hypothetical (e.g., "it's not involving the central part of the macula, but he can certainly progress"), DO NOT discard the visual fact! Extract ONLY the verbatim substring of the physical finding (e.g., "it's not involving the central part of the macula") for the 'scenes'. The hypothetical remainder of that sentence should go to 'supplemental\_context'.

- 4. DISCARD to 'supplemental\_context': All hypotheticals, prognoses, general pathology definitions, historical filler, and lecture logistics.

\textbf{OUTPUT FORMAT:}
You must output a valid JSON object. Separate distinct images/views into different objects within the "scenes" array.

\{
  "supplemental\_context": [
    "<Verbatim list of phrases containing general medical context, hypotheticals, or prognoses>"
  ],
  "scene\_count": <integer representing how many distinct images the speaker refers to>,
  "scenes": [
    {
      "scene\_id": 1,
      "verbatim\_scene\_text": "<The exact, unmodified phrase/sentence from the transcript containing BOTH the spatial/visual cue and the physical finding. STRICTLY NO PROGNOSES.>"
    }
  ]
\}

\end{promptbox}

\subsection{Implementation of Visual cue separation and scoring.}
Since raw audio transcripts often contain conversational fillers and general medical background that do not directly correspond to the visual content, we prompt an LLM judge (i.e., Qwen3.5-9B), to decouple each transcript into two components: \texttt{Scenes} and \texttt{Supplemental Context}. This step ensures that the generated data is grounded in explicit visual features rather than hallucinated from the model's prior knowledge. Acting as a strict data extractor, the judge partitions the content without modification or paraphrasing. The system prompt is shown in Fig.~\ref{prompt:system-visualcue}. We also provide a one-shot example for in-context guidance. During inference, we set the temperature to 0, use a maximum generation length of 2048 tokens, and disable Qwen's thinking mode.

\begin{promptbox}[label=prompt:system-visualscore]{System prompt for scoring.}
You are an expert data evaluator specializing in multimodal ophthalmology datasets, specifically retinal imaging (e.g., OCT, fundus photography, fluorescein angiography). Your task is to evaluate text extracted from clinical transcripts. 

Note: You will NOT be provided with the actual images. You must evaluate how well the text alone serves as a proxy for the visual content and provides clear medical context.

The input JSON payload contains:

- 1. "supplemental\_context": The surrounding video transcript. This provides the broader background discussion and medical context, but will also include natural speaker transitions (e.g., "Now let's look at..."), lecture mechanics, or conversational filler.

- 2. "scenes": Extracted visual cues and verbatim narration describing the unseen image. Note that the JSON formatting here may be imperfect, malformed, or grouped irregularly. Do not penalize the text for JSON syntax errors.

Evaluate the provided JSON text payload based purely on its intrinsic quality, clinical complexity, and descriptive power. You must first provide a brief (3-4 sentences) reasoning for your evaluation, and then output your scores. Output your results strictly in JSON format. Do not include any explanation, conversational text, or markdown formatting outside of the JSON block.

\textbf{EVALUATION CRITERIA:}

- 1. Quality (Score 1-10) Assess the clarity, accuracy, and descriptive power of the text. 

9-10: Extremely clear, accurate, and highly descriptive. Perfectly articulates clinical concepts or what is physically happening in a patient case with no ambiguity.

7-8: Clear and accurate expression, but may have minimal ambiguity or slightly less descriptive detail.

5-6: Fairly clear and generally accurate, but some notable ambiguity exists. Contains disjointed thoughts or conversational filler alongside clinical data.

3-4: Not very clear, somewhat vague expression, with obvious ambiguity or heavily fragmented text.

1-2: Unclear, very vague expression, difficult to understand, or entirely conversational filler with no descriptive value.

- 2. Difficulty (Score 1-10) Assess the clinical complexity of the text and the level of specialized medical knowledge required to understand it.

9-10: Highly complex. Requires expert-level ophthalmology knowledge (e.g., analyzing laser wavelength absorption in the RPE vs. hemoglobin, complex surgical mechanics, specific biomarker interactions).

7-8: Quite difficult. Requires specialized clinical knowledge and analysis (e.g., identifying specific stages of macular edema, disease progression, or specific laser applications).

5-6: Moderate difficulty. Requires general medical knowledge or basic eye anatomy.

3-4: Fairly simple. Can be understood with basic, entry-level knowledge.

1-2: Very simple or generic statements. No special medical knowledge required.

- 3. Relevance2Medicine (Score 1-6) Assess the clinical relevance of the text to ophthalmology.

5-6: Highly relevant. Packed with dense medical terminology, specific treatments, or imaging jargon (e.g., "retinal pigment epithelium", "microaneurysms", "PRP").

3-4: Moderately relevant. Involves general medical fields or basic eye anatomy, but lacks deep specialized terminology.

1-2: Very weak or no medical relevance.

- 4. MentionSpecificDetails (Boolean: true/false) Determine if specific case details, anatomical landmarks, or precise patient presentations are mentioned.

true: Mentions explicit details (e.g., "scotomas involving the periphery", "retina nasal to the nerve", specific lesion locations, or exact measurements). Do not penalize for a lack of patient identifiers or exact numerical measurements.

false: Generalized statements, abstract concepts without concrete examples, or lacking pinpointed details about a case.

- OUTPUT FORMAT:Strictly output a valid JSON object using the exact keys below.

\{
  "reasoning": "<Brief 3-4 sentence explanation of why these scores were chosen>",
  "quality": <Integer 1-10>,
  "difficulty": <Integer 1-10>,
  "Relevance2Medicine": <Integer 1-6>,
  "MentionSpecificDetails": <Boolean true or false>
\}
\end{promptbox}

Following visual cue separation, we prompt a second LLM judge (i.e., Qwen3.5-9B), to assess transcript quality under clinical expert guidance. Its system prompt is shown in Fig.~\ref{prompt:system-visualscore}. Given the \texttt{Scenes} and \texttt{Supplemental Context} arrays, the judge evaluates each transcript using four criteria and provides supporting rationales: \texttt{Quality} on a 1 to 10 scale, \texttt{Difficulty} on a 1 to 10 scale, \texttt{Relevance2Medicine} on a 1 to 6 scale, and the boolean field \texttt{MentionSpecificDetails}. Specifically, \texttt{Quality} measures textual clarity and descriptive richness; \texttt{Difficulty} reflects clinical complexity; \texttt{Relevance2Medicine} assesses relevance to ophthalmology; and \texttt{MentionSpecificDetails} verifies whether the transcript contains case-specific details or anatomical landmarks that can serve as explicit visual cues. We only retain instances that are ophthalmology-relevant (i.e., \texttt{Relevance2Medicine} $\geq$ 3) and contain explicit visual cues (i.e., \texttt{MentionSpecificDetails} is true). We provide three-shot in-context examples for additional guidance. During inference, we set the temperature to 0, use a maximum generation length of 2048 tokens, and disable the thinking mode.

\subsection{Implementation of Instruction Synthesis}

\begin{promptbox}[label=prompt:system-vqa]{System prompt for simple VQA generation.}
You are an expert ophthalmic data engineer building a Visual Question Answering (VQA) dataset. You will be provided with a JSON payload representing a medical lecture about an unseen image. 

The input JSON payload contains:

- 1. "supplemental\_context": The surrounding video transcript. This provides the broader background discussion and medical context, but will also include natural speaker transitions (e.g., "Now let's look at..."), lecture mechanics, or conversational filler.

- 2. "scenes": Extracted visual cues and verbatim narration describing the unseen image. Note that the JSON formatting here may be imperfect, malformed, or grouped irregularly. 

\textbf{YOUR CORE TASK \& DATA HIERARCHY:}

Your absolute source of truth is the "scenes" array. Act as if you are looking directly at the physical elements described in these scenes. You may use the "supplemental\_context" ONLY as a passive medical dictionary to resolve ambiguous terms or find the formal clinical names for the visual features explicitly shown in the scenes. NEVER use the supplemental context to ask about overall diagnoses not explicitly visible, patient history, or conversational filler.

\textbf{CRITICAL RULES (The "Pixel Test"):}

- 1. Visual Grounding Only: The question must be answerable purely by pointing to pixels on a screen. If a detail cannot be physically seen, do not ask a question about it.

- 2. No Transcript Framing: Never mention the text, the speaker, the lecture, or the video. Ask the question directly as if holding the image (e.g., "What is visible in the retina?").

- 3. Extreme Conciseness: VQA answers must be incredibly short—often just 1 to 5 words. Do not write full sentences for the answers.

- 4. The Anti-Hallucination Escape Hatch: If the assigned "{TARGET\_TYPE}" cannot be answered using ONLY the visual facts explicitly stated in the scenes (for example, you are assigned a "where" question but the text contains zero spatial information), DO NOT hallucinate. Instead, output "N/A" for both the question and answer.

\textbf{TARGET QUESTION TYPE:}

Based strictly on the visual facts in the "scenes", you must generate exactly ONE "{TARGET\_TYPE}" question. 

- If "yes\_no": Ask a binary question confirming the presence, absence, or state of a specific visual feature. The answer must be exactly "Yes." or "No."

- If "what": Ask a question identifying a specific pathology, treatment mark, or anatomical structure.

- If "where": Ask a spatial question localizing a specific finding within the image (e.g., quadrant, relative position, or anatomical zone).

\textbf{OUTPUT FORMAT:}
Strictly output ONLY a valid JSON object. Do not wrap the JSON in markdown formatting or code blocks (e.g., no json). Start your response directly with the opening curly brace.

\{
  "reasoning": "<String: Briefly explain which exact phrase from the 'scenes' you are using, why it passes the Pixel Test, and how it fits the assigned question type. If triggering the Escape Hatch, explain why the text lacks the necessary information.>",
  "question\_type": "{TARGET\_TYPE}",
  "question": "<String or 'N/A'>",
  "answer": "<String or 'N/A'>"
\}
\end{promptbox}

\noindent \textbf{Simple VQA} This subset is designed to improve the model’s ophthalmology-specific visual comprehension by ensuring that responses are explicitly grounded in observed image features rather than hallucinated from prior knowledge. Specifically, we use an LLM generator (i.e., Qwen3.5-27B) to formulate question-answer pairs based exclusively on visual cues in the \texttt{Scenes} component. As shown in the system prompt in Fig.~\ref{prompt:system-vqa}, the generator is required to produce auxiliary reasoning chains to justify its outputs and is strictly prohibited from referencing the raw transcripts. To improve linguistic diversity, we require three question types for each instance: \texttt{Yes/No}, \texttt{What}, and \texttt{Where}. To further mitigate hallucination, the model is instructed to output \texttt{N/A} when the visual metadata is insufficient to support a factual QA pair. Throughout this process, \texttt{Supplemental Context} serves only as a passive medical reference for clarifying ambiguous clinical terms. We do not provide additional few-shot examples and use the same generation configuration as in the previous stage. Here, we disable the thinking mode of Qwen.

\begin{promptbox}[label=prompt:system-conv]{System prompt for conversation data generation.}

You are an AI assistant specialized in medical and ophthalmic imaging. You are provided with a direct visual description ([SCENE]) of a medical image from an expert transcript. You also have additional clinical text ([SUPPLEMENTAL CONTEXT]) that discusses related disease mechanisms, stages, or treatments. Unfortunately, you don't have access to the actual image.

Your task is to generate a conversation between a person (User) inquiring about the image and you (Assistant) responding to their questions. The conversation should proceed as though both the User and Assistant are actively viewing the image together, while never explicitly referring to the provided text information ([SCENE] and [SUPPLEMENTAL CONTEXT]).

Below are the strict requirements for generating the questions and answers in the conversation:

- DO NOT have the User ask to "analyze a text" or "summarize a description." The User must ask natural, conversational questions as if pointing at the image. 

- The first question must focus on the visual aspects of the image (drawn from [SCENE]). Subsequent questions should explore deeper clinical understanding, mechanisms, or treatments related to what is seen (drawn from [SUPPLEMENTAL CONTEXT]).

- You must use the specific medical facts, terms, and diagnoses from the provided text to make the conversation highly educational. However, present them as clinical observations or expert knowledge rather than reading from a script.

- Do not use phrases like "the text mentions", "in the scene", "the context states", or "based on the clinical description." Instead, refer to the findings as being "in the image," "in this scan," or "visible here."

- Ensure that questions are diverse and flow logically from visual findings to clinical implications.

- The conversation should include 3 to 4 turns of back-and-forth questions and answers.

- Answer responsibly, avoiding overconfidence, and do not provide medical advice or diagnostic information. Encourage the user to consult a healthcare professional for advice.

- Output the conversation STRICTLY as a raw JSON array of objects with "from" (either "user" or "assistant") and "value" (the text), without any markdown blocks.

\end{promptbox}

\noindent \textbf{Conversation Interactions.} To enhance the model’s capacity for expert-level clinical dialogue, we introduce an additional LLM generator (Qwen3.5-27B) to synthesize a multi-turn conversational subset. As shown in the system prompt in Fig.~\ref{prompt:system-conv}, the generator constructs each dialogue as if the participants are actively examining the provided image. To ensure strict visual grounding, the opening turn is formulated exclusively from visual cues in the \texttt{Scenes} metadata. Unlike standard VQA data, subsequent turns may incorporate information from \texttt{Supplemental Context} to support deeper clinical discussion of disease mechanisms, staging, or treatment options. We provide few-shot examples, use the same generation configuration as in the previous stage, and disable Qwen’s thinking mode.

\begin{promptbox}[label=prompt:system-cot]{System prompt for CoT reasoning generation.}
You are an expert medical AI data generator specializing in ophthalmology, neuro-ophthalmology, and comprehensive medical imaging. Your task is to generate high-quality, single-turn conversational data for Supervised Fine-Tuning (SFT). This data is strictly for academic AI research and study. It will be used to train a multimodal AI model to reason step-by-step about diverse medical images to facilitate deep understanding and clinical evaluation before delivering a final answer.

\textbf{YOUR INPUT:}

You will be provided with a formatted text block containing two main sections extracted from case presentations:

- 1. [SUPPLEMENTAL CONTEXT]: A bulleted list (e.g., "- [fact]") providing background information, medical reasoning, surgical steps, and sometimes irrelevant conversational filler (e.g., "- I mean, let me show you...").

- 2. [SCENE]: Itemized content (e.g., "Scene 1: [quote]") containing verbatim direct quotes that refer to visual findings on a medical image.

\textbf{CRITICAL CONSTRAINT:}

You CANNOT see the actual images. You must rely purely on the provided text to deduce what is happening visually. 

\textbf{RULES FOR GENERATION:}

- 1. The User (`user` value):

- Formulate a realistic question from the perspective of a medical professional.

- Assume the image is already the shared context between the user and the assistant. The user should ask direct, natural questions without explicitly stating they are looking at an image (e.g., avoid "Looking at this scan..."). 

- Seamlessly integrate the visual finding from the [SCENE] text into the question (e.g., "Why are there significant scotomas involving the periphery?").

- 2. The Assistant (`assistant` value):

- Provide a response that follows a clear, logical progression of thought before delivering the final answer.

- Absolute Roleplay Constraint: You MUST act as if you are directly looking at the image alongside the user. NEVER use words like "text," "transcript," "context," "quote," "scene," or "audio." Do not mention that you cannot see the image.
   
- Flexible Structured Reasoning: You must construct your response using a consistent cognitive sequence, but you must blend these steps naturally. Do NOT use rigid, robotic labels (like "First," "Second," "Third," or "Therefore"). 

- The Fallback Rule: If the [SUPPLEMENTAL CONTEXT] is irrelevant, empty, or lacks sufficient medical facts, you MUST rely entirely on your own expert medical knowledge to explain the visual findings.

- Ensure your response flows through these three logical phases:
     
- Observation Phase: Acknowledge and detail the specific visual findings on the image that are relevant to the user's query. Internally scan the [SCENE] data to deduce these visual findings and describe them as if you are seeing them.

- Correlation Phase: Connect these visual findings to the underlying pathophysiology, anatomy, or procedural mechanics. Internally scan the [SUPPLEMENTAL CONTEXT] bullets for relevant medical clues and apply them.
     
- Conclusion Phase: Synthesize the reasoning into a final, direct clinical explanation that answers the user's question.
   
- Use natural, varied transition phrases (e.g., "Based on these visual features...", "This indicates that...", "Clinically, this correlates with...") to clearly separate your phases of thought while sounding like an expert physician. Ensure your transitions make anatomical sense and do not blindly copy these examples.

\textbf{OUTPUT FORMAT:}

You must output ONLY valid JSON using the exact schema below. It must be an array containing two objects. Do not include markdown formatting or conversational text outside the JSON block.

{\ttfamily
[\\
\quad \{\\
\quad\quad "from": "user",\\
\quad\quad "value": "<A realistic, direct question asking about the visual findings>"\\
\quad \},\\
\quad \{\\
\quad\quad "from": "assistant",\\
\quad\quad "value": "<A single, concise paragraph (roughly 3--5 sentences) smoothly blending the Observation, Correlation, and Conclusion phases. Ground the reasoning in the visual reality and transition naturally into a final academic/clinical answer.>"\\
\quad \}\\
]
}

\end{promptbox}

\noindent \textbf{CoT Reasoning} For this subset, we prompt an LLM generator (i.e., GPT-4.1) to synthesize single-turn CoT data. We select only high-quality instances satisfying \texttt{Quality} $\geq 9$, \texttt{Difficulty} $\geq 9$, \texttt{Relevance2Medicine} $\geq 5$, and \texttt{MentionSpecificDetails} $=$ true. As shown in Fig.~\ref{prompt:system-cot}, each question is structured into a three-phase reasoning chain: (1) an observation phase that describes specific visual findings; (2) a correlation phase that links these findings to underlying clinical mechanisms; and (3) a conclusion phase that provides the final answer with a clinical explanation. When the transcript lacks sufficient detail, the generator is allowed to use its internal medical knowledge to bridge these logical transitions. We provide few-shot examples as conversational context, set the maximum generation length to 2048 tokens, and use a temperature of 0.4 to encourage richer reasoning exploration.

\begin{promptbox}[label=prompt:system-vqa-check]{System Prompt for VQA quality checking.}
You are an expert medical AI data evaluator. Your task is to evaluate single-turn Visual Question Answering (VQA) data for multimodal AI training in ophthalmology.

IMPORTANT CONSTRAINT:You CANNOT see the actual image. You MUST rely purely on the provided text to evaluate the quality of the VQA pair.

YOUR INPUT:You will receive a text block with three sections:

- 1. [SUPPLEMENTAL CONTEXT]: A bulleted list (e.g., "- [fact]") providing background information, medical reasoning, surgical steps, and sometimes irrelevant conversational filler (e.g., "- I mean, let me show you...").

- 2. [SCENE]: Itemized content (e.g., "Scene 1: [quote]") containing verbatim direct quotes that refer to visual findings on a medical image.

- 3. [VQA CONTENT]: The generated question and answer formatted as a JSON conversation array.

EVALUATION CRITERIA:

- 1. The question MUST ONLY target visual elements in the [SCENE]. Use [SUPPLEMENTAL CONTEXT] ONLY as a medical dictionary to resolve terms or find formal clinical names for those specific visuals.

- 2. The Q\&A MUST simulate a user looking directly at an image. Discard any pair that references the text (e.g., "According to the text," "As mentioned").

- 3. The answer MUST be derivable from the provided inputs. Discard any hallucinations or invented clinical findings.

OUTPUT FORMAT: Output ONLY a valid JSON object. Do NOT wrap the JSON in markdown code blocks (e.g., do not use json or ). Ensure your reasoning outlines the evaluation against the four criteria before providing the final answer.

\{
"reasoning": "<String: Step-by-step evaluation against the three criteria to justify the final decision.>",
"answer": "<String: strictly 'keep' or 'discard'>"
\}

\end{promptbox}

\begin{promptbox}[label=prompt:system-conv-check]{System Prompt for conversation quality checking.}
You are an expert medical AI data evaluator. Your task is to evaluate multi-turn conversation data for multimodal AI training in ophthalmology.

IMPORTANT CONSTRAINT: You CANNOT see the actual image. You MUST rely purely on the provided text to evaluate the quality of the conversation data.

YOUR INPUT: You will receive a text block with three sections:

- 1. [SUPPLEMENTAL CONTEXT]: A bulleted list (e.g., "- [fact]") providing background information, medical reasoning, surgical steps, and sometimes irrelevant conversational filler (e.g., "- I mean, let me show you...").

- 2. [SCENE]: Itemized content (e.g., "Scene 1: [quote]") containing verbatim direct quotes that refer to visual findings on a medical image.

- 3. [CONVERSATION]: The generated questions and answers formatted as a JSON conversation array.

EVALUATION CRITERIA:

- 1. The conversation MUST start by discussing the visual findings explicitly mentioned in the [SCENE]. From there, it should flow naturally and may expand into broader, relevant medical topics.

- 2. The dialogue MUST simulate a natural clinical conversation about an image. Discard any conversation that breaks character by referencing the source text (e.g., "According to the transcript," "The speaker mentioned," "Based on the text").

- 3. Specific case facts (e.g., patient history, specific visual findings, exact treatments given) MUST come strictly from the provided inputs. Discard any conversation that invents or hallucinates case-specific details.

- 4. It is ACCEPTABLE and ENCOURAGED for the conversation to incorporate general medical knowledge (e.g., identifying drug classes, explaining mechanisms of action, or discussing general side effects) to enrich the dialogue, provided this external knowledge does NOT alter or contradict the actual case facts.

OUTPUT FORMAT: Output ONLY a valid JSON object. Do NOT wrap the JSON in markdown code blocks (e.g., do not use json or ). Ensure your reasoning outlines the evaluation against the four criteria before providing the final answer.

\{
"reasoning": "<String: Step-by-step evaluation against the four criteria to justify the final decision.>",
"answer": "<String: strictly 'keep' or 'discard'>"
\}

\end{promptbox}

\begin{promptbox}[label=prompt:system-cot-check]{System Prompt for CoT quality checking.}
You are an expert medical AI data evaluator. Your task is to evaluate single-turn Chain-of-Thought (CoT) data for multimodal AI training in ophthalmology. This data is strictly for academic AI research and study. 

IMPORTANT CONSTRAINT:You CANNOT see the actual image. You MUST rely purely on the provided text to evaluate the quality of the CoT data.

YOUR INPUT:You will receive a text block with three sections:

- 1. [SUPPLEMENTAL CONTEXT]: A bulleted list (e.g., "- [fact]") providing background information, medical reasoning, surgical steps, and sometimes irrelevant conversational filler (e.g., "- I mean, let me show you...").

- 2. [SCENE]: Itemized content (e.g., "Scene 1: [quote]") containing verbatim direct quotes that refer to visual findings on a medical image.

- 3. [GPT CoT]: The generated CoT data formatted as a JSON conversation array.

EVALUATION CRITERIA:

- 1. The dialogue MUST simulate a natural clinical conversation about an image. Discard any conversation that breaks character by referencing the source text (e.g., "According to the transcript," "The speaker mentioned," "Based on the text").

- 2. The user's question AND the assistant's initial response MUST be anchored in the visual findings provided in the [SCENE].

- 3. The transitions of assistant's response MUST be natural and conversational. Discard responses that give a flat, immediate answer without reasoning, AND discard responses that use robotic, rigid formatting (e.g., "Observation Phase:", "First:", "Conclusion:"). 

- 4. It is ACCEPTABLE and ENCOURAGED for the assistant to introduce external medical knowledge to bridge logical gaps in its reasoning. Discard conversations ONLY if this external knowledge alters or directly contradicts the specific case facts (e.g., patient history, specific visual findings, exact treatments given) found in the provided texts.

OUTPUT FORMAT: Output ONLY a valid JSON object. Do NOT wrap the JSON in markdown code blocks (e.g., do not use json or ). Ensure your reasoning outlines the evaluation against the four criteria before providing the final answer.

\{
"reasoning": "<String: Step-by-step evaluation against the four criteria to justify the final decision.>",
"answer": "<String: strictly 'keep' or 'discard'>"
\}

\end{promptbox}

\subsection{Implementation of Post-Processing and Quality Control}
Following data synthesis, the engine applies a final LLM-based verification stage to filter failed generations, including ill-posed questions, hallucinated content, and persona-breaking responses. The verification prompts for VQA, conversational interactions, and CoT reasoning are shown in Fig.~\ref{prompt:system-vqa-check}, Fig.~\ref{prompt:system-conv-check}, and Fig.~\ref{prompt:system-cot-check}, respectively. We use Qwen3.5-27B as the verifier for VQA and conversational data, and GPT-4.1 for CoT reasoning data. For the simple VQA subset, we first discard unanswerable queries (i.e., those labeled as \texttt{N/A}). The verifier then checks whether each QA pair is strictly grounded in visual evidence, without relying on transcripts or introducing fabricated details. For conversational data, in addition to enforcing visual grounding in the opening turn and maintaining persona consistency throughout the dialogue, we require the verifier to reject hallucinated case-specific facts, such as fictitious patient histories or unobservable visual findings. However, general medical knowledge, such as common side effects, is allowed when it enriches the interaction without contradicting the visual case. For CoT reasoning data, we apply similar grounding criteria and further discard instances that lack intermediate logical steps or exhibit overly rigid, templated formatting. We retain only instances explicitly assigned a \texttt{keep} decision by the verifier. After automated filtering, clinical experts help to conduct sample-based verification to provide an additional quality assessment. During verification, we set the temperature to 0, use a maximum generation length of 2048 tokens, and disable Qwen’s thinking mode.

\section{Extended Analysis of OphIn-500K}
\begin{table}[htbp]
\centering
\resizebox{0.95\columnwidth}{!}{
\begin{tabular}{@{} l l r r r | r r r @{}}
\toprule
\multirow{2}{*}{\textbf{Data type}} 
& \multirow{2}{*}{\textbf{Subtype}} 
& \multirow{2}{*}{\textbf{Instances}} 
& \multirow{2}{*}{\textbf{Train}} 
& \multirow{2}{*}{\textbf{Test}} 
& \multicolumn{3}{c}{\textbf{Unique images by modality}} \\
\cmidrule(lr){6-8}
& & & & & \textbf{CFP} & \textbf{OCT} & \textbf{UWF} \\
\midrule

\multirow{3}{*}{VQA} 
& Yes/No & 145,252 & 144,802 & 450 & 37,738 & 76,207 & 31,307 \\
& What   & 142,971 & 142,520 & 451 & 37,202 & 74,914 & 30,855 \\
& Where  & 110,898 & 110,565 & 333 & 28,817 & 58,221 & 23,860 \\
\midrule

Conversation 
& \textit{N/A} & 124,441 & 124,441 & -- & 32,006 & 65,412 & 27,023 \\

CoT 
& \textit{N/A} & 12,570 & 12,570 & -- & 2,745 & 6,883 & 2,942 \\
\midrule

\textbf{Total} 
& \textit{N/A} & \textbf{536,132} & \textbf{534,898} & \textbf{1,234} 
& \textbf{38,943} & \textbf{80,139} & \textbf{32,348} \\
\midrule

\multicolumn{2}{@{}l}{\textbf{Source videos}} 
& \multicolumn{2}{r}{\textbf{29,465 clips}} 
& \multicolumn{4}{r@{}}{\textbf{$\sim$14,700 hours}} \\

\bottomrule
\end{tabular}
}
\caption{Statistics of OphIn-500K. The dataset is constructed from 29,465 video clips totaling approximately 14,700 hours. The VQA subset is further split into training and test sets for model development and evaluation.}
\label{tab:sft-datastatis}
\end{table}

\begin{figure*}[ht]
    \centering
    \includegraphics[width=0.9\linewidth]{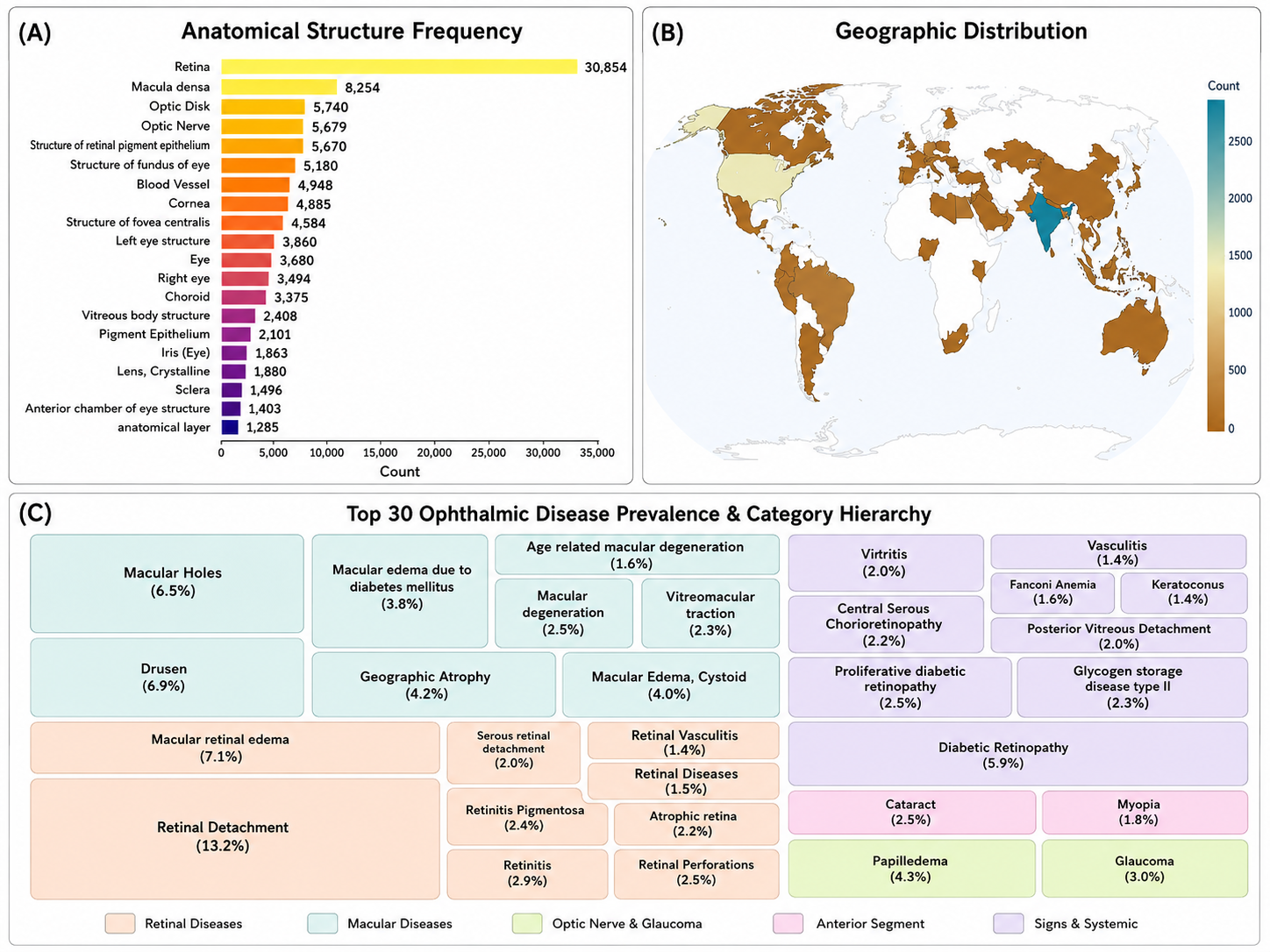}
    \caption{Illustration of OphIn-500K statistics. \textbf{(A)} represents the top 20 retinal anatomical structures. \textbf{(B)} denotes its geographic distribution across the words and \textbf{(C)} shows the top 30 ophthalmic disease prevalence and category hierarchy. All statistics are derived from the corresponding raw audio transcriptions.}
    \label{fig:final-data-statistics}
\end{figure*}

Processed through OphIn-Engine, OphIn-500K contains 536,132 instruction instances from 151,430 unique images, extracted from 29,465 video clips totaling approximately 14,700 hours. The dataset covers three major ophthalmic imaging modalities, including CFP, OCT, and UWF, and spans diverse clinical concepts, including over 1,000 retinal conditions and approximately 800 anatomical features across 100 countries and regions. Additional visualizations of major retinal anatomical structures, global geographic distribution, and ophthalmic disease prevalence and hierarchy are provided in Fig.~\ref{fig:final-data-statistics}. Unlike prior ophthalmology instruction datasets that largely rely on existing public benchmarks, OphIn-500K is constructed from real-world web video sources, providing visually novel image-text pairs at scale. To support rigorous evaluation, we further construct an OphIn-VQA test split by sampling representative VQA instances across major diseases, anatomical structures, and question types. The final test split contains 1,234 instances, including 450 Yes/No, 451 What, and 333 Where questions, while the remaining VQA instances are retained for training. Detailed dataset statistics are provided in Tab.~\ref{tab:sft-datastatis}.

\section{Implementation Details of OphIn-VL}
\subsection{Training Implementation Details.}

\begin{promptbox}[label=prompt:system-training]{System Prompt for OphIn-VL training.}
1. \textbf{VQA} 
- You are an expert ophthalmology AI assistant. Analyze the provided clinical image and answer the question as concretely and concisely as possible. For yes/no questions, respond strictly with "Yes." or "No." For "what" or "where" questions, provide only the specific anatomical structure, finding, or condition, without any additional explanation.

2. \textbf{Conversation Interactions}
- You are an expert ophthalmology AI assistant. We are having a multi-turn clinical discussion. Analyze the provided clinical images and address the queries accurately and professionally, utilizing the context of our ongoing conversation to provide clinical insights.

3. \textbf{CoT Reasoning} 
- You are an expert ophthalmology AI assistant. Analyze the provided clinical imaging and provide a comprehensive assessment. Structure your response through three logical phases: First, detail the direct visual observations relevant to the query. Second, correlate these visual findings with the underlying pathophysiology, anatomical structures, or procedural mechanics. Finally, synthesize your reasoning into a direct clinical conclusion that answers the query.
\end{promptbox}

OphIn-VL adopts Qwen-3.5-9B~\cite{qwen3.5} as its core language backbone and uses the SWIFT~\cite{zhao2024swiftascalablelightweightinfrastructure} training pipeline for efficient multimodal supervised fine-tuning. The vision encoder is kept frozen throughout training to preserve pretrained visual representations and reduce memory and computational cost. We apply LoRA~\cite{hu2022lora} to all linear modules in both the vision-language projector and language backbone, with rank $r=32$ and scaling factor $\alpha=64$. The model is trained for 2 epochs using AdamW with an initial learning rate of $2\times10^{-4}$, a cosine learning-rate scheduler, a maximum sequence length of 4096 tokens, and a per-device batch size of 4. Training is performed on 4 NVIDIA A100 80GB GPUs with DeepSpeed ZeRO-2~\cite{rajbhandari2020zero} and bfloat16 (bf16) mixed precision. During SFT, we compute the cross-entropy loss only on model responses, while masking system prompts and user queries. Unless otherwise specified, all remaining optimization, precision, checkpointing, dataloader, and auxiliary training configurations follow the default settings of the SWIFT framework. The complete fine-tuning process takes approximately 2 days on 4 NVIDIA A100 80GB GPUs, resulting in an estimated compute cost of 192 A100 GPU-hours. Detailed training configurations are summarized in Tab.~\ref{tab:ophinvl_training_config}.

OphIn-500K contains three instruction formats: VQA, conversational interaction, and CoT reasoning. Therefore, we use format-specific system prompts to guide different response behaviors. As illustrated in Fig.~\ref{prompt:system-training}, the VQA prompt encourages concise and grounded answers for ophthalmic image understanding; the CoT prompt guides the model to follow a structured reasoning process from observation and correlation to final conclusion; and the conversational prompt encourages clinically informative multi-turn interaction. OphIn-VL is trained on all instances in the OphIn-500K training split, which contains 534,898 instruction instances. These include 124,441 general multi-turn conversational interactions, 12,570 chain-of-thought reasoning examples, and categorized VQA samples. The VQA subset consists of 144,802 Yes/No, 142,520 What, and 110,565 Where instances. By combining answer-oriented supervision with conversational and reasoning-based instruction formats, OphIn-VL learns both accurate visual recognition and ophthalmology-specific interactive reasoning.

\begin{table}[t]
\centering
\caption{Training configuration of OphIn-VL. Unless otherwise specified, all remaining hyperparameters follow the default settings of the SWIFT framework.}
\label{tab:ophinvl_training_config}
\resizebox{\linewidth}{!}{
\begin{tabular}{ll}
\toprule
\textbf{Configuration} & \textbf{Setting} \\
\midrule
Base language backbone & Qwen-3.5-9B \\
Training framework & SWIFT \\
Training objective & Multimodal supervised fine-tuning \\
Vision encoder & Frozen \\
Trainable components & Vision-language projector and language backbone \\
Parameter-efficient tuning & LoRA \\
LoRA target modules & All linear modules in trainable components \\
LoRA rank $r$ & 32 \\
LoRA scaling factor $\alpha$ & 64 \\
Loss computation & Cross-entropy loss on model responses only \\
Input masking & System prompts and user queries are masked \\
System prompts & Format-specific prompts for VQA, conversation, and CoT data \\
Number of epochs & 2 \\
Optimizer & AdamW \\
Initial learning rate & $2 \times 10^{-4}$ \\
Learning-rate scheduler & Cosine scheduler \\
Maximum sequence length & 4096 tokens \\
Per-device batch size & 4 \\
Distributed training & DeepSpeed ZeRO-2 \\
Training hardware & 4 $\times$ NVIDIA A100 80GB GPUs \\
Training time & Approximately 2 days \\
Estimated compute cost & Approximately 192 A100 GPU-hours \\
Total training instances & 534,898 \\
Instruction formats & VQA, multi-turn conversation, and CoT reasoning \\
Conversational instances & 124,441 \\
CoT reasoning instances & 12,570 \\
Yes/No VQA instances & 144,802 \\
What VQA instances & 142,520 \\
Where VQA instances & 110,565 \\
Other hyperparameters & SWIFT default settings \\
\bottomrule
\end{tabular}
}
\end{table}
\subsection{Evaluation Implementation Details.}

\begin{promptbox}[label=prompt:system-evaluation-judge]{System Prompt for LLM judge in OphIn-VQA}
You are an impartial, strict expert in ophthalmology, neuro-ophthalmology, and comprehensive medical imaging. Your task is to evaluate the factual correctness of a model's answer to a medical question.

\textbf{YOUR INPUT}:

You will be provided with a formatted text block containing five main sections extracted from case presentations:

- 1. \textbf{[SUPPLEMENTAL CONTEXT]}: A bulleted list (e.g., "- [fact]") providing background information, medical reasoning, surgical steps, and sometimes irrelevant conversational filler (e.g., "- I mean, let me show you...").

- 2. \textbf{[SCENE]}: Itemized content (e.g., "Scene 1: [quote]") containing verbatim direct quotes that refer to visual findings on a medical image.

- 3. \textbf{[Question]}: A question synthesized based on information in the [SCENE].

- 4. \textbf{[LABEL]}: The ground truth answer. This is the absolute source of truth for your evaluation. Use the Context and Scene only to understand the clinical scenario.

- 5. \textbf{[RESPONSE]}: The model response, which is the target you need to evaluate.

\textbf{CRITICAL CONSTRAINT:}

- 1. You CANNOT see the actual images. You must rely purely on the provided text to evaluate the performance.

- 2. DO NOT give preference to conversational, wordy, or detailed responses.

- 3. A concise, short answer MUST receive a 1.0 if it captures the core facts of the ground truth.

- 4. DO NOT penalize for grammatical incompleteness.

- 5. If the model refuses to answer (e.g., "Refusal"), score it as 0.0.

\textbf{RULES FOR SCORING:}

You must score the model's answer on a scale from 0.0 to 1.0 using the following strict rubric:

    - 1.0: The answer is factually correct, complete, and perfectly aligns with the ground truth.
    
    - 0.75: The answer is mostly correct and relevant but is missing a very minor detail.
    
    - 0.5: The answer is partially correct but misses major parts of the ground truth or includes some irrelevant info.
    
    - 0.25: The answer is mostly incorrect but contains a tiny sliver of relevant truth.
    
    - 0.0: The answer is completely incorrect, irrelevant, or contradicts the ground truth.

\textbf{OUTPUT FORMAT:}

You must output ONLY valid JSON using the exact schema below. Output a single JSON object (not an array). Do not include markdown formatting or conversational text outside the JSON block.

\{
  ``reasoning'': ``<String: Briefly explain your medical reasoning for the score>'',
  ``score'': <Number: 1.0, 0.75, 0.5, 0.25, or 0.0>
\}
\end{promptbox}

We evaluate all models on the OphIn-VQA split, which contains 1,234 open-ended VQA test instances: 450 \texttt{Yes/No}, 451 \texttt{What}, and 333 \texttt{Where} questions. The corresponding images are held out from the OphIn-500K training split to ensure fair evaluation. Since all questions are open-ended, we adopt an LLM-as-a-judge protocol to assess answer accuracy, with the system prompt defined in Fig.~\ref{prompt:system-evaluation-judge}. We also report the F1 score of BERTScore~\cite{zhang2019bertscore} to measure semantic similarity between model predictions and reference answers. To reduce evaluation bias caused by different response styles, such as verbose explanations or concise formatted answers, we first use an auxiliary LLM extractor to normalize each model response into a concise answer while preserving its factual content. The extractor prompt is provided in Fig.~\ref{prompt:system-evaluation-extractor}. Both the LLM-as-a-judge score and BERTScore are computed on the normalized outputs. We use GPT-4.1-mini as both the judge and extractor agent.

\begin{promptbox}[label=prompt:system-evaluation-extractor]{System Prompt for LLM extractor}
You are a strict data-extraction assistant. Your only job is to extract the final, core factual answer from the model's response based on the original question.

    - You must output ONLY the extracted concise answer.
    
    - DO NOT include conversational filler (e.g., 'The image shows...', 'The answer is...').
    
    - DO NOT use punctuation unless it is part of the answer itself.
    
    - If the response implies the model cannot answer the question (e.g., I cannot answer the question), output exactly: 'Refusal'."
\end{promptbox}

For OphIn-VL generation on the OphIn-VQA split, we use greedy decoding by setting the temperature to 0 and the maximum number of new tokens to 2048 across all experiments and case-study illustrations. We compare OphIn-VL with a broad set of open-source MLLM baselines, including general medical, ophthalmology-specific, and general-domain models. The general medical baselines include LLaVA-Med~\cite{li2023llava}, MedGemma~\cite{sellergren2025medgemma}, Lingshu~\cite{xu2025lingshu}, and MedVLM-R1~\cite{pan2025medvlm}; the ophthalmology-specific baselines include RetinalGPT~\cite{zhu2025retinalgpt}, FundusExperts~\cite{liu2025constructing}, and OphthaReason~\cite{wu2025bridging}; and the general-domain baselines include Qwen~\cite{qwen3.5}, LLaMA~\cite{grattafiori2024llama}, and InternVL~\cite{chen2024internvl}. For all baselines, we keep their default inference settings except that we set the temperature to 0 and the maximum number of new tokens to 2048 for consistency. For RL fine-tuned models, such as MedVLM-R1 and the OphthaReason series, we use only the content enclosed by \texttt{<answer>} and \texttt{</answer>} as the final response, while retaining the reasoning traces only for qualitative case studies. The detailed checkpoint versions are listed in Tab.~\ref{tab:evaluation_baselines}. 

\begin{table}[t]
\centering
\caption{Baseline selection for OphIn-VQA evaluation. \textbf{Default} denotes the official released checkpoint when a method provides a fine-tuned model based on a single default backbone.}
\label{tab:evaluation_baselines}
\resizebox{0.8\linewidth}{!}{
\begin{tabular}{lll}
\toprule
\textbf{Category} & \textbf{Model} & \textbf{Version / Checkpoint} \\
\midrule
\multirow{4}{*}{General medical}
& LLaVA-Med~\cite{li2023llava} & llava-med-v1.5-mistral-7B \\
& MedGemma~\cite{sellergren2025medgemma} & MedGemma-1.5-4B-it \\
& Lingshu~\cite{xu2025lingshu} & Lingshu-7B-i1-GGUF \\
& MedVLM-R1~\cite{pan2025medvlm} & Default \\
\midrule
\multirow{3}{*}{Ophthalmology-specific}
& RetinalGPT~\cite{zhu2025retinalgpt} & Default \\
& OphthaReason~\cite{wu2025bridging} & OphthaReason-Qwen/Intern \\
& FundusExperts~\cite{liu2025constructing} & Default \\
\midrule
\multirow{3}{*}{General-domain}
& Qwen~\cite{qwen3.5} & Qwen-3.5-9B \\
& LLaMA~\cite{grattafiori2024llama} & LLaMa-3.2-11B-vision \\
& InternVL~\cite{chen2024internvl} & InternVL3-8B \\
\bottomrule
\end{tabular}
}
\end{table}

\end{document}